\documentclass{article}
\usepackage[utf8]{inputenc}

\usepackage{times}
\usepackage{epsfig}
\usepackage{graphicx}
\usepackage{amsmath}
\usepackage{amssymb}
\usepackage{enumerate}
\usepackage{xcolor}
\usepackage{caption}
\usepackage{subcaption}
\usepackage[round,sort]{natbib}

\newcommand{\comment}[1]{}

\title{Artwork Identification from Wearable Camera Images for Enhancing Experience of Museum Audiences}
\author{Rui Zhang, Yusuf Tas, Piotr Koniusz}
\date{April 22, 2017\thanks{This work has been published at MW17: Museums and the Web 2017.}}

\begin{document}
\maketitle

\begin{abstract}
Recommendation systems based on image recognition could prove a vital tool in enhancing the experience of museum audiences. However, for practical systems utilizing wearable cameras, a number of challenges exist which affect the quality of image recognition. In this pilot study, we focus on recognition of museum collections by using a wearable camera in three different museum spaces. We discuss the application of wearable cameras, and the practical and technical challenges in devising a robust system that can recognize artworks viewed by the visitors to create a detailed record of their visit. Specifically, to illustrate the impact of different kinds of museum spaces on image recognition, we collect three training datasets of museum exhibits containing variety of paintings, clocks, and sculptures. Subsequently, we equip selected visitors with wearable cameras to capture artworks viewed by them as they stroll along exhibitions. We use Convolutional Neural Networks (CNN) which are pre-trained on the ImageNet dataset and fine-tuned on each of the training sets for the purpose of artwork identification. In the testing stage, we use CNNs to identify artworks captured by the visitors with a wearable camera. We analyze the accuracy of their recognition and provide an insight into the applicability of such a system to further engage audiences with museum exhibitions.
\end{abstract}

\section{Introduction}

A vast number of approaches exist dedicated to engaging and educating audiences in museums, e.g. augmented reality, mobile guides, interactive collections and 3D displays, to name a few. Artworks in museums engage visitors with their past experiences and trigger effective response which constitutes a vital aspect of a positive museum experience~\citep{alelis_emotional}. The value of emotional experiences in museums has been linked to reinforced trust, increased chances of recurring visits, as well as gaining donations~\citep{suchy}. 

However, the experience of visitors is often incomplete because of the limited space dedicated to museum exhibitions, and personal time constraints during the visit. \cite{beer} pointed out that museum visitors spend less than one minute with each artwork during a typical visit. To a large extent, the audience has a limited idea of artworks they want to view or topics they are excited to cover. Therefore, they visit museums based on personal recommendations, advertisements, or a rough idea of the topics a museum covers. Viewers often adopt a fast pace as they stroll along through exhibition space, giving an incomplete or repetitive experience. Moreover, museums and cultural sites often lack interactive or personalized entertainment gadgets, guideline systems, and other technology to customize visits efficiently~\citep{baraldi_gesture}.

It is undeniable that museum audiences have access to smart phones and virtual interactive technology. However, robust guide systems that help satisfy their expectations and enhance their emotional experience are still rare. \cite{kuflik} proposed a system customizing user’s experience which employs statistical machine learning capable of inferring visitors’ interests, based on their answers to a pre-specified questionnaire. By analogy, in order to aid a museum curator’s work, wearable or security cameras could provide an input to autonomous software which in turns would perform an analysis of audiences’ preferences inside the museum. Such a system could count numbers of visitors, capture the time they spend with specific artworks, or even attempt to recognize their mood based on facial expressions, in order to isolate the most popular artworks, as well as consider visitor’s likes and dislikes. However, wearable devices have limited processing power, and memories which are based on so-called local feature descriptors~\citep{dalens_painting}. Nonetheless, more robust end-to-end recognition systems such as Convolutional Neural Networks (CNN) have been shown to be particularly well suited for object category recognition~\citep{krizhevsky_alexnet}. We therefore assess the suitability of CNNs for image recognition of museum artworks captured with wearable cameras. CNNs require a lot of computational resources at their training stage. However, they can perform real-time recognition on Android-based systems with a camera.

In this work, we use wearable cameras for capturing images of artworks captured `in-the-wild' by audiences as they stroll along three different museum spaces and interact with various artworks. We used the data we collected to study the ability of CNNs to identify specific artworks in images. 

As artworks vary from paintings, to sculptures, to other unusual rigid and non-rigid shapes and texture forms, we illustrate the impact of different types of museum spaces on image recognition. Specifically, we first collect non-occluded images of art pieces in each exhibition space with a phone camera. Next, we use the database of images collected by the audiences as they stroll with wearable cameras for testing recognition accuracy.

In the training stage, we use CNN pre-trained on the ImageNet dataset~\citep{ILSVRC15} and fine-tune such a pre-trained CNN on each of our datasets for the purpose of artwork identification. Due to the major technical challenges in image recognition such as non-planar sculptures, glare of protective cabinets, reflective properties of surfaces, background clutter, occlusions, rotations, scale changes, viewpoint changes, lighting variations, motion blur, and other limiting factors, this work is conducted as a pilot study to identify the impact of these phenomena on recognition. The results will provide a better understanding of whether a wearable camera-based system can be used to help audiences engage with museum exhibitions, and if they reliably identified artworks from wearable cameras that could be used as an input for a recommendation system.

\section{Artwork identification with wearable cameras}

Our work aims to identify artworks using wearable cameras in the context of the museum. Our hope is that museums might benefit from wearable technology in order to improve guidance and management of audiences. For this purpose, we choose three different types of museum spaces that pose varied challenges in terms of image capturing with wearable cameras.

\vspace{0.05cm}
\noindent\textbf{Shenzhen Museum of Arts}, located in Shenzhen, Guangdong Province, China, has a diverse collection of artworks such as traditional Chinese paintings, oil paintings, prints, sculptures, calligraphy, watercolors, caricatures, paper-cuttings, and photographic works. For this study, we capture the Chinese traditional paintings from this museum.

\vspace{0.05cm}
\noindent\textbf{The Palace Museum}, located in Beijing, China, is a home to the Clock and Watch Gallery as well as the Indian and Chinese Sculpture Exhibition (AD400-700). The collections in the Clock and Watch Gallery consist of more than two hundred clocks from the 18th century. The sculptures of the Indian and Chinese Sculpture Exhibition mainly include Buddhist statues from India and China from AD400 to AD700.

\begin{figure}[t]
\centering
\comment{
\begin{subfigure}[b]{0.105\linewidth}
\centering\includegraphics[trim=0 0 0 0, clip=true, height=2.3cm]{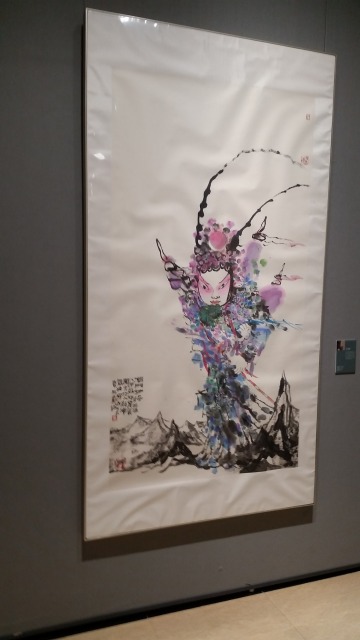}
\end{subfigure}
\begin{subfigure}[b]{0.105\linewidth}
\centering\includegraphics[trim=0 0 0 0, clip=true, height=2.3cm]{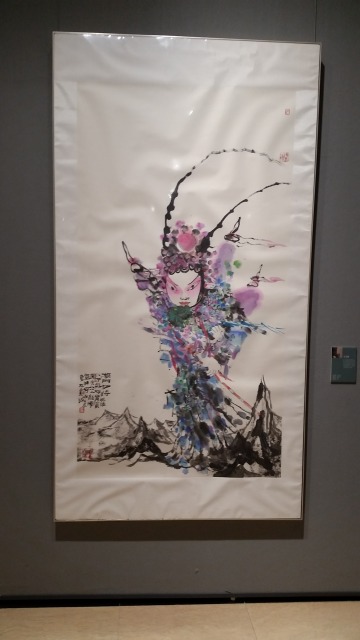}
\end{subfigure}
\begin{subfigure}[b]{0.105\linewidth}
\centering\includegraphics[trim=0 0 0 0, clip=true, height=2.3cm]{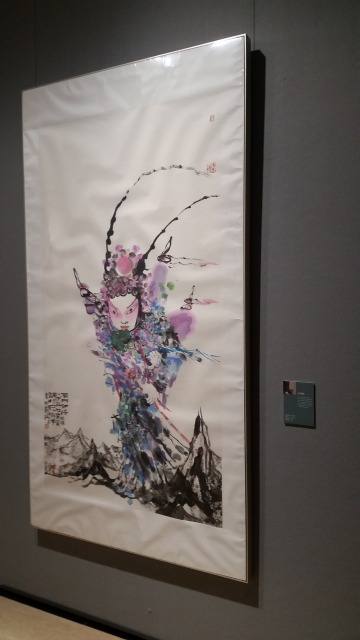}
\end{subfigure}
\begin{subfigure}[b]{0.105\linewidth}
\centering\includegraphics[trim=0 0 0 0, clip=true, height=2.3cm]{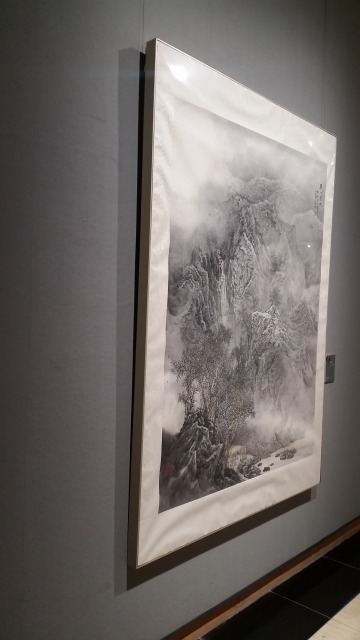}
\end{subfigure}
\begin{subfigure}[b]{0.105\linewidth}
\centering\includegraphics[trim=0 0 0 0, clip=true, height=2.3cm]{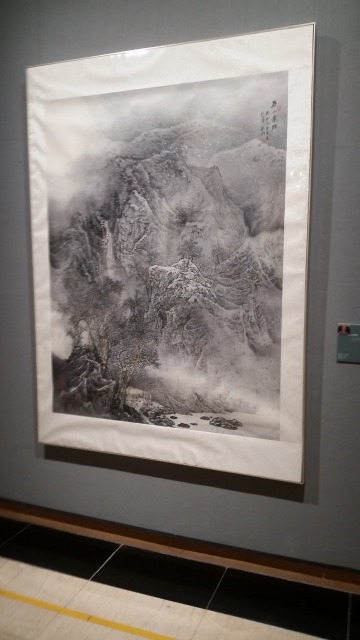}
\end{subfigure}
\begin{subfigure}[b]{0.105\linewidth}
\centering\includegraphics[trim=0 0 0 0, clip=true, height=2.3cm]{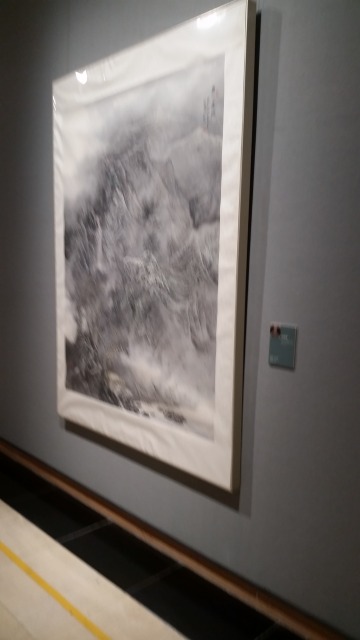}
\end{subfigure}
\begin{subfigure}[b]{0.105\linewidth}
\centering\includegraphics[trim=0 0 0 0, clip=true, height=2.3cm]{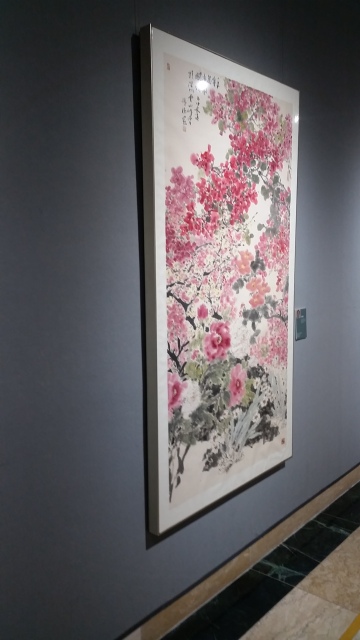}
\end{subfigure}
\begin{subfigure}[b]{0.105\linewidth}
\centering\includegraphics[trim=0 0 0 0, clip=true, height=2.3cm]{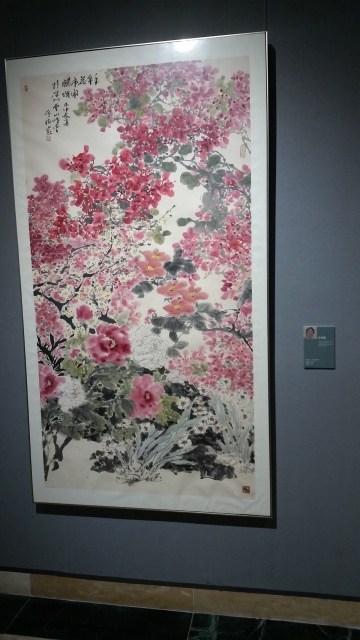}
\end{subfigure}
\begin{subfigure}[b]{0.105\linewidth}
\centering\includegraphics[trim=0 0 0 0, clip=true, height=2.3cm]{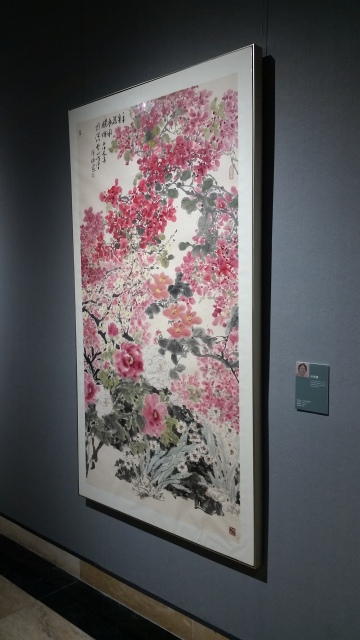}
\end{subfigure}
}
\includegraphics[trim=0 0 0 0, clip=true, height=2.325cm]{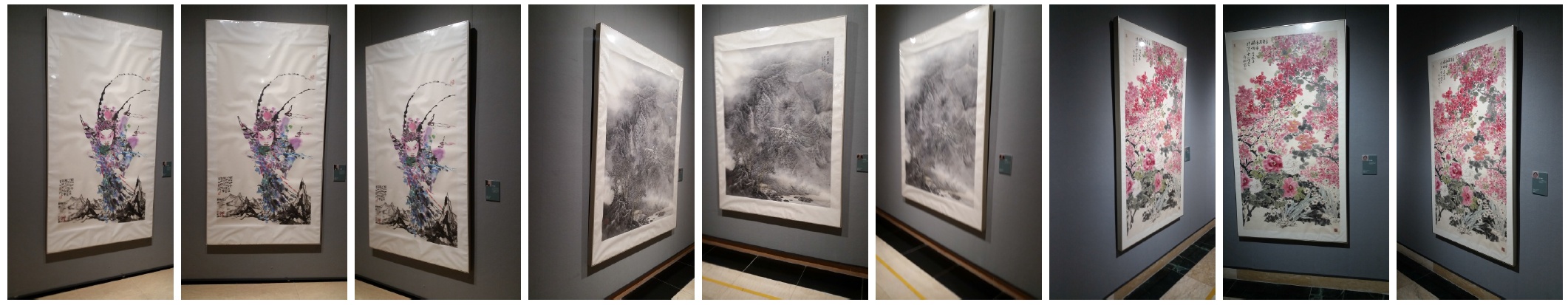}
%
\caption{Examples of Shenzhen paintings. For the training set, we captured paintings from various viewpoints.}
\label{fig:paintings}
\end{figure}

\begin{figure}[t]
\centering
\hspace*{-0.2cm}
\includegraphics[trim=0 0 0 0, clip=true, height=6.8cm]{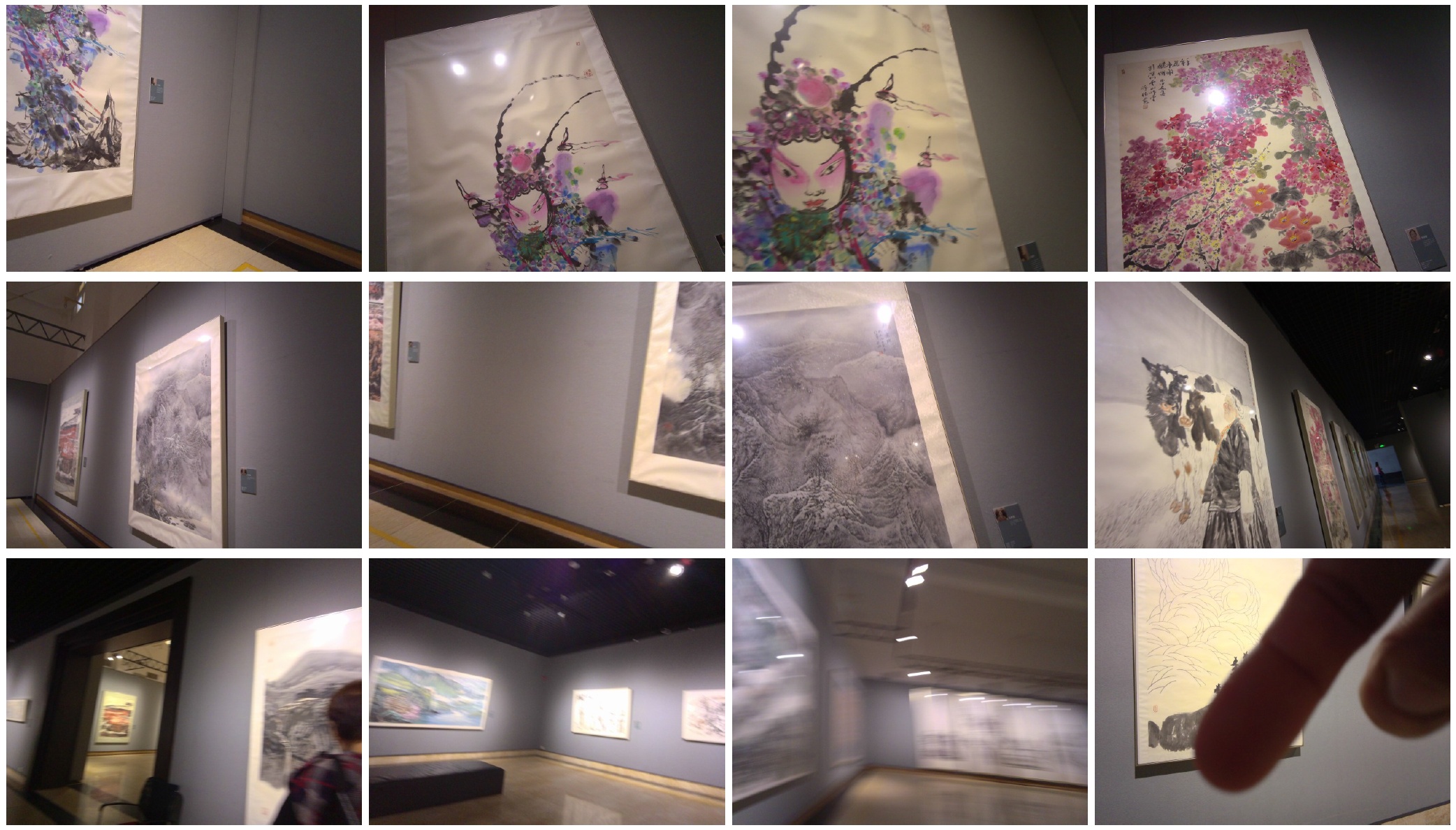}
\caption{Shenzhen paintings. Top and middle rows show that the geometric transformations resulting from capturing the test set by the wearable camera are large. They include perspective changes, zoom, rotations and cropping. The bottom row also shows an occlusion by person, glare,  motion blur and an occlusion by hand.}\vspace{-0.2cm}
\label{fig:paintings2}
\end{figure}

\subsection{Data collection}
\label{sec:data_coll}
In order to train a recognition algorithm, we needed to collect a dataset of “objects to identify.” For this purpose, we used an ordinary Android phone. To account for viewpoint and scale changes, we captured between two and six photos of each artwork viewed from different viewpoints and distances. For testing purposes, we equiped six volunteers with a wearable camera and asked them to walk the exhibition space and interact with artworks. Afterwards, we annotated these images with labels assigned to the artworks that can be seen in each image. The wearable camera is configured to capture a picture every 10 seconds.

\vspace{0.05cm}
\noindent\textbf{Shenzhen Paintings} consists of 79 distinct paintings that were displayed in the museum during the capturing process, each photographed several times, resulting in the total of 369 images. Figure 1 illustrates that these paintings were captured under several viewpoints. We also included a background category representing museum surroundings, which consists of 27 images, and a spurious category of 170 miscellaneous paintings that were not on display. The latter subset helps to refine the classifier which has to distinguish between the 79 specific instances of paintings, other possible artworks, and the background. This resulted in 566 training images. For the testing set, we equipped six volunteers with the wearable camera and collected six different splits, as detailed below.

Split 1 contains 86 images from the wearable camera, which was mounted at the right-hand side pocket at upper chest height. Split 2 contains 93 images from the camera mounted on the right-hand side of a jacket zipped up to chest height. Split 3 contains 54 images from the partially rotated camera mounted on the left-hand side belt of a backpack at the mid-chest height. Split 4 includes 86 close-up images from the camera mounted on the collar. Splits 5 and 6 contain 91 and 105 images from the camera mounted on a handbag strap at chest height and left-hand side bottom, respectively.

\begin{figure}[t]
\centering
\includegraphics[trim=0 0 0 0, clip=true, height=4.65cm]{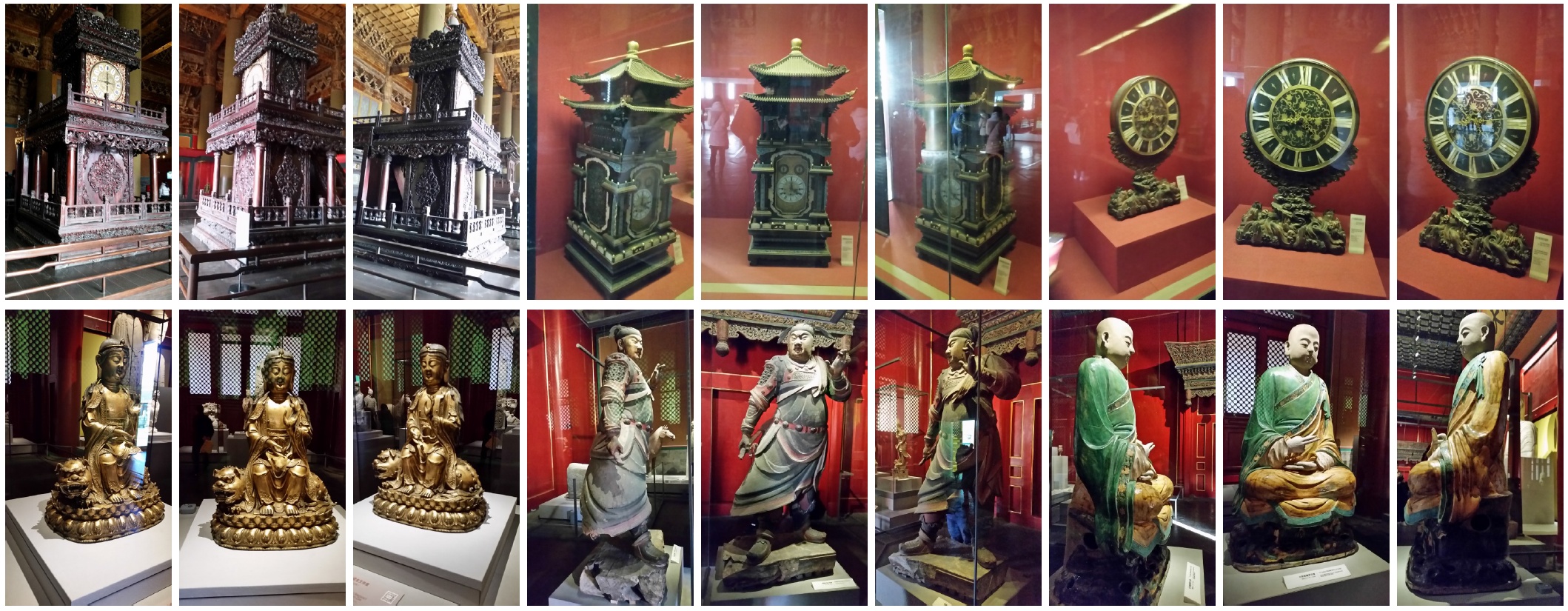}
%
\caption{Examples of pieces from the Clock and Sculptures training sets are given in the top and bottom row, respectively. Note the non-planarity of these pieces as well as glares from the protective glass.}\vspace{-0.45cm}
\label{fig:clocks_train}
\end{figure}

\begin{figure}[!b]
\centering\vspace{-0.2cm}
\includegraphics[trim=0 0 0 0, clip=true, height=3.05cm]{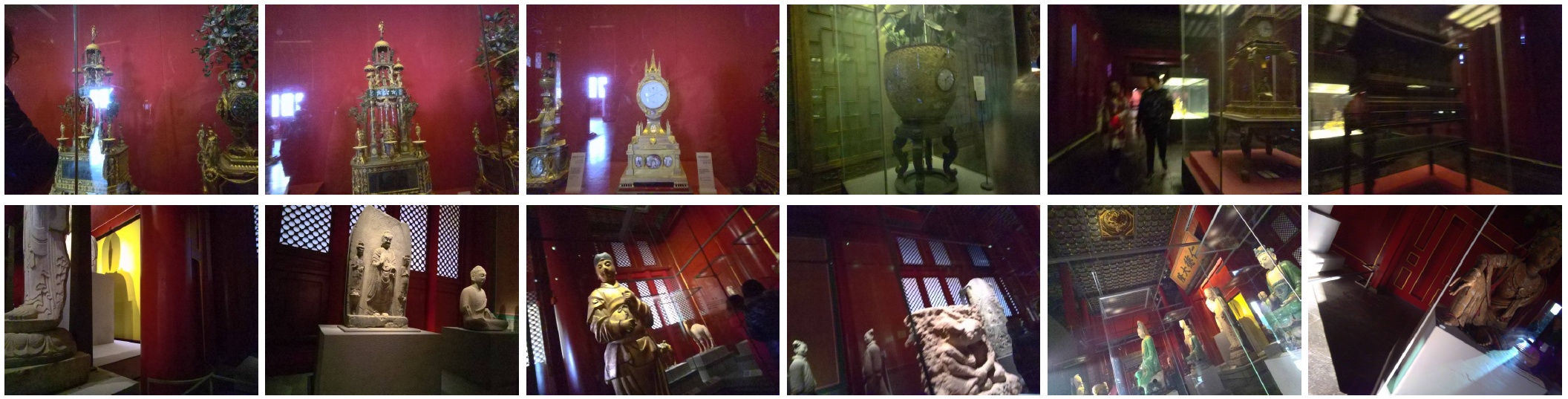}
\caption{Examples from the Clock and Sculptures testing sets are given in the top and bottom row, respectively. Glares, viewpoint changes, rotations, background clutter, occlusions, salt and pepper noise \citep{gonzalez_dip_book} occur in large quantities.}\vspace{-0.45cm}
\label{fig:clocks_test}
\end{figure}

Figure \ref{fig:paintings2} illustrates images captured by the wearable camera and resulting transformations which make recognition a challenge. In total, the testing set resulted in 515 images of paintings. We annotated each image with ground truth labels that indicate the paintings which are visible in these images (ordered from the most visible artwork to the least visible one). During the training stage, we chose one of the splits for testing and the remaining five splits for validation. Therefore, to obtain accuracy on all six splits, we had to repeat the training six times. To enhance our study by recognizing artworks other than paintings, which are planar, we collected the following datasets:

\vspace{0.05cm}
\noindent\textbf{The Clocks} dataset consists of 113 distinct clocks, each photographed several times, resulting in 394 images. Additionally, we captured 27 images of backgrounds not containing any clocks. For validation, we captured a separate set with the Android camera, which contains 259 images. Lastly, for testing, we devised two splits captured by two volunteers consisting of 182 and 141 images. They were captured with a camera mounted on the pocket (the top of chest) and on the handbag belt (mid-chest) with straight and rotated orientations, respectively. Overall, this resulted in 653 training and 323 testing images. Examples of clocks from training and testing sets are shown in Figure \ref{fig:clocks_train} (top) and Figure \ref{fig:clocks_test} (top).

\vspace{0.05cm}
\noindent\textbf{The Sculptures} dataset consists of 44 distinct sculptures, each photographed several times, resulting in 206 images. An additional two categories were created which consist of photos of sculpture descriptions which may contain only tiny fragments of sculptures and 27 images of background. Two testing splits were captured by volunteers and resulted in 80 and 50 images, respectively. The cameras were mounted on the handbag belt (mid-chest) with clockwise and counterclockwise orientations, respectively. When testing on the first split, the second one is used for validation, and vice versa. Overall, training and testing sets resulted in 233 and 130 images, respectively. While this is the smallest testing set, it is also the most challenging, due to large nonplanar sculptures on display in several locations, which include other sculptures in the background. Examples of sculptures from training and testing sets are shown in Figure \ref{fig:clocks_train} (bottom) and Figure \ref{fig:clocks_test} (bottom).

\subsection{Image recognition}
\begin{figure}[t]
\centering
\begin{subfigure}[b]{0.99\linewidth}
\centering\includegraphics[trim=0 0 0 0, clip=true, width=10.5cm]{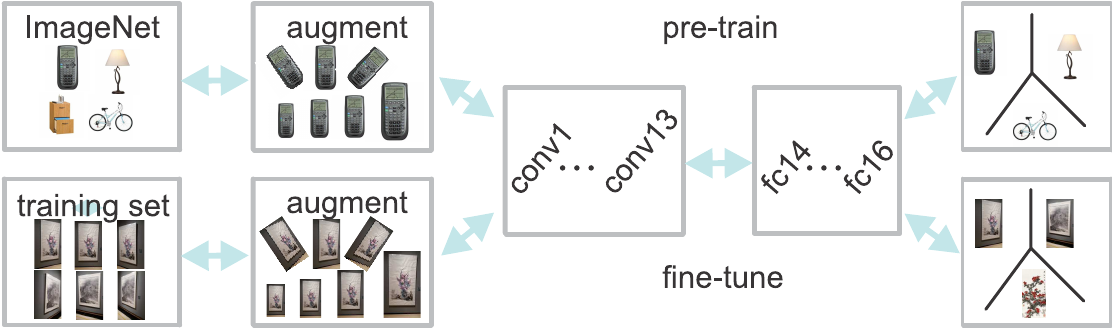}
\caption{\label{fig:train}}
\end{subfigure}
\begin{subfigure}[b]{0.99\linewidth}
\centering\includegraphics[trim=0 0 0 0, clip=true, width=10.5cm]{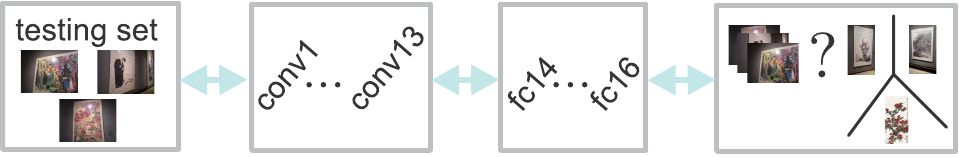}
\caption{\label{fig:test}}
\end{subfigure}
%
\caption{Training of CNN. In Figure \ref{fig:train}, the network is shown to be first pre-trained on augmented images from the ImageNet dataset. Then, augmented training set is used for so-called fine-tuning to adapt the network to recognize the training set. Figure \ref{fig:test} shows the testing stage.}
\label{fig:train_test}
\end{figure}
For the purposes of artwork identification, we employ one of the latest CNN architectures known as VGG16~\citep{vgg16} which consists of 13 so-called convolutional layers and three fully connected layers which results in an extremely large number of network parameters that need to be inferred in the training stage. Therefore, we pre-train it with the ImageNet dataset containing over 14 million images and 1000 object categories. Subsequently, we utilize the training data we collected to perform image augmentations~\citep{krizhevsky_alexnet} and we fine-tune the VGG16 network on these images. Details and discussions on fine-tuning can be found in numerous literature~\citep{fine_tune}. TThe hyper-parameters are selected in the cross-validation process by used validation sets, as described in Section \ref{sec:data_coll}. Lastly, in the testing stage, we applied the trained network to our test sets in a feed-forward manner, quantifying whether identification agrees with the ground truth. This outcome is indicative of whether CNN can reliably recognize what visitors see in wearable camera images. Figure \ref{fig:train_test} illustrates the pipeline used in our experiments.

\vspace{0.05cm}
\noindent\textbf{Data augmentation.} A standard technique to train CNN representations, which are somewhat invariant to partial image translation, rotation, scale and viewpoint changes, is to augment the training dataset with multiple crops of images (e.g. left, right, top, bottom, center crops), mirroring images by left-right flips, arbitrary rotations, and contrast changes. We apply this technique to each training set to replicate expected variations between training and testing splits resulting from the capturing process.

\section{The impact of different types of museum spaces on data capturing and image recognition}
In the museum context, design decisions such as interior light, visitor circulation quality, audiences’ time limitations, layout of showcases, size of artwork and type of artwork have been shown to affect the number of pieces that visitors will encounter. Without a doubt, these factors will also impact the quality of images captured by wearable cameras. For instance, good quality uniform illumination will be positively correlated with the acquisition of crisp images. However, artworks with scarce lighting will result in images that show signs of the sensor noise, e.g. the salt-and-pepper noise known in digital photography \citep{gonzalez_dip_book}.

\subsection{Paintings}
\label{sec:mus_paintings}
Paintings in Shenzhen Art Museum are displayed in an ordered manner within the given exhibition space. The arrangement of paintings in the exhibition halls influences to what degree audiences interact with these paintings, and it imposes some natural order in which artworks are viewed and captured by our wearable cameras. 

\begin{figure}[t]
\centering
\begin{subfigure}[b]{0.42\linewidth}
\centering\includegraphics[trim=0 0 0 0, clip=true, width=5.0cm]{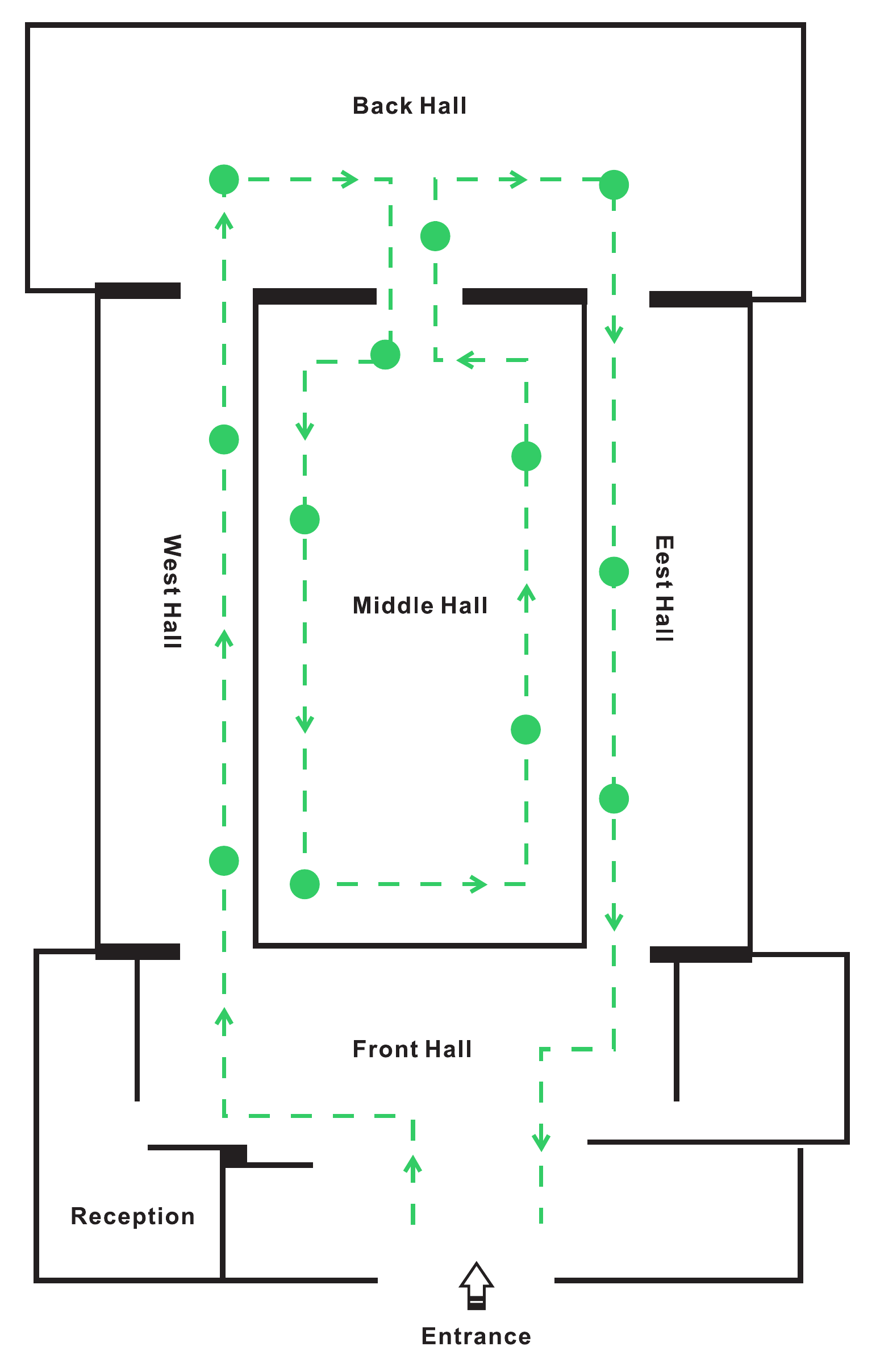}
\caption{\label{fig:museum1}}
\end{subfigure}
\begin{subfigure}[b]{0.57\linewidth}
\centering\includegraphics[trim=0 0 0 0, clip=true, width=7.25cm]{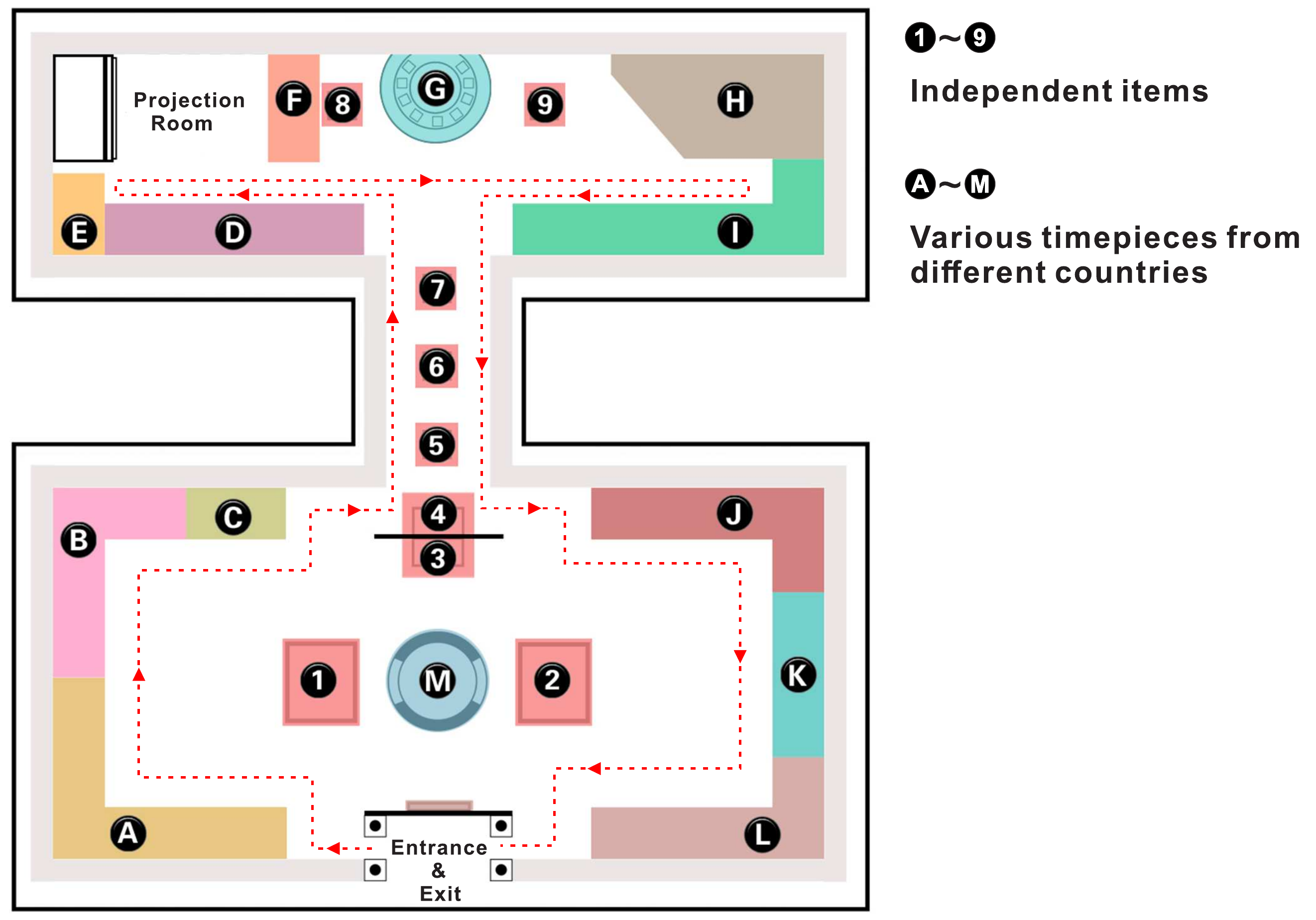}
\caption{\label{fig:museum2}}
\end{subfigure}
%
\caption{The layouts of Shenzhen Art Museum and the Palace Museum are given in Figure \ref{fig:museum1} and \ref{fig:museum2}, respectively.}
\label{fig:museums}
\end{figure}

Specifically, paintings that are located close to each other on the wall are usually captured in one shot. Therefore, pictures from wearable cameras often contain more than one paining per image. While it may be hard to determine which paintings in an image are of direct interest to visitors, some paintings are captured more than once, thus increasing a chance of successful identification. Moreover, the number of captures of the same painting potentially correlates with the time and interest dedicated to such a painting. Additionally, some paintings are captured partially, e.g. they are truncated, making recognition even harder. Other practical issues include shadows cast on artworks due to lighting and proximity of the painting to the viewer, as well as motion blur.
The layout of the museum imposes some partial order in which paintings are displayed and captured by wearable cameras. According to the path delineated in Figure \ref{fig:museum1}, we see that the visitor’s route is clear and easy to follow, so that audiences are not likely to miss many artworks. Volunteers who walked around the exhibition wearing cameras tended to stop next to various paintings for various durations of time. Moreover, they could easily avoid revisiting the same artworks, unless they desired to approach some of them again. Note that audiences in this museum are not allowed to touch any artworks, however, they can look at any of the paintings close-up, which may result in a partial capture, i.e. zooming in at a fragment of a painting. The spacious exhibition hall provided good conditions for our volunteers to capture images in a varied manner; some viewers preferred to approach artworks, others just strolled along at a steady pace. Therefore, we were able to collect six testing sets, as detailed in Section \ref{sec:data_coll}. 

\subsection{Clocks}
\label{sec:mus_clocks}
The clocks in the Palace Museum are displayed in cases, under the necessary preservation conditions. Because they are located behind glass, the clocks cannot usually be interacted with through touch or seen from extremely close up; and because it is difficult to get clear shots of artworks due to the low-light interior environment and reflections from the glass surfaces, the photos may be blurry due to overexposure. Moreover, we also noted that it was hard to take shots from acute viewpoints. This complication was due to clocks being located close to each other; taking photos of the rear side of these clocks was often impossible, as only their frontal parts were clearly exposed to the visitors. Of course, this specific constraint on viewpoints seems to have a positive effect in the sense that it limits the number of views an object can be seen from, while the front view remains very distinct. However, three clocks are displayed without any glass case due to their large size, and can be seen by viewers from all sides. These artworks are still protected from audiences by handrails. In this case, the artworks stretch beyond the field of view of the wearable camera, making it difficult to capture good images of entire objects. This is especially undesirable, because if only partial views are being captured, the representations of these artworks are much less distinct.

The Palace Museum is a very popular attraction, with large numbers of tourists visiting everyday. This crowded space resulted in some photos of timepieces that were partially occluded by visitors. Therefore, adverse conditions described above differ from the case outlined in Section \ref{sec:mus_paintings} and should affect, to some degree, identification of the artworks. Lastly, the red dotted line in figure 6b illustrates visitor circulation in the Clock and Watch Gallery.

\subsection{Sculptures}
In the Indian and Chinese Sculpture Exhibition Hall, many sculptures are located in the middle of the exhibition space and therefore they are set against background cluttered by other sculptures. This makes both the annotation of ground-truth data and its identification a challenging process as numerous art pieces are often captured at once. 
The way audiences move in this museum space has more complex pattern compared to the case study in Section \ref{sec:mus_clocks}. Volunteers often exhibited counterclockwise movement around the perimeter if they turned right at the entrance and clockwise movement otherwise. In the hall area with artworks located on both sides, volunteers often followed a zigzag path between these art pieces. Moreover, volunteers also often lapped circles around the smaller patches emerging between art pieces. Therefore, in this dataset, one cannot expect a clear order in which art pieces were captured nor a clear correlation between frequently viewed sculptures and audiences' preference. Another adverse factors included large-sized sculptures which did not fit well into the field of view of wearable camera, occlusions, and poor lighting. In our opinion, such factors make this exhibition space the most challenging for the purpose of capturing images with wearable cameras.

\subsection{Other exhibition spaces}
During the capturing process in the three different art museum spaces, we observed that viewers are actively enjoying the parts of exhibitions they are interested in, while ignoring the others. Using this process, artworks can be identified using scientific tools that give museums the opportunity to re-think the way they communicate, i.e. beyond offering the standard guided tours and fixed exhibitions \citep{balsamo}. However, each museum space poses unique challenges for artwork identification. For instance, science museum exhibitions often include items which are large and may look very similar to the untrained eye, such as engines, pumps, radio-communication equipment, etc. These items may be bulky, highly non-planar, and not clearly localized; they may emit light, change appearance during interaction, and so forth. Other artworks such as crafts are also likely to be highly non-planar, e.g. miniature replicas of houses, famous buildings, and monuments. Exhibitions with non-rigid objects such as carpet, Gobelin tapestry, and clothing are a further example of artworks of varied nature in the exhibition spaces. Modern art may include objects that lack texture, making them harder to recognize, while porcelain and glass work are likely to be the source of glares. Hieroglyphs, ancient books, and even jewelry may all look similar to a non-expert eye. Exhibits in natural history museums such as birds, insects, butterflies, rodents, etc. may pose similar challenges. These last two are examples of so-called fine grained image recognition \citep{fine_grained} which requires an algorithm match and expert knowledge about what makes these exhibits differ between many similar items. However, we leave these challenges for future work.

\section{Experiments}
To conclude our work, below we present experimental findings from our study. We separately fine-tuned three VGG16 networks for the paintings, clocks and sculptures, respectively. To achieve this, we followed the augmentation and cross-validation process as detailed in Section \ref{sec:data_coll}. Below, we report results in terms of mean accuracy, which quantifies how many test images on average were assigned labels agreeing with our ground truth annotations. Note that some images annotated by us contained more than one museum item. We assigned ground truth labels to these images in descending order; that is, the central artwork was assigned its ground truth label first while less visible peripheral pieces were assigned their ground truth labels next.

\begin{figure}[t]
\centering\vspace*{-0.6cm}
\begin{subfigure}[b]{0.49\linewidth}
\centering\includegraphics[trim=0 0 0 0, clip=true, width=6.0cm]{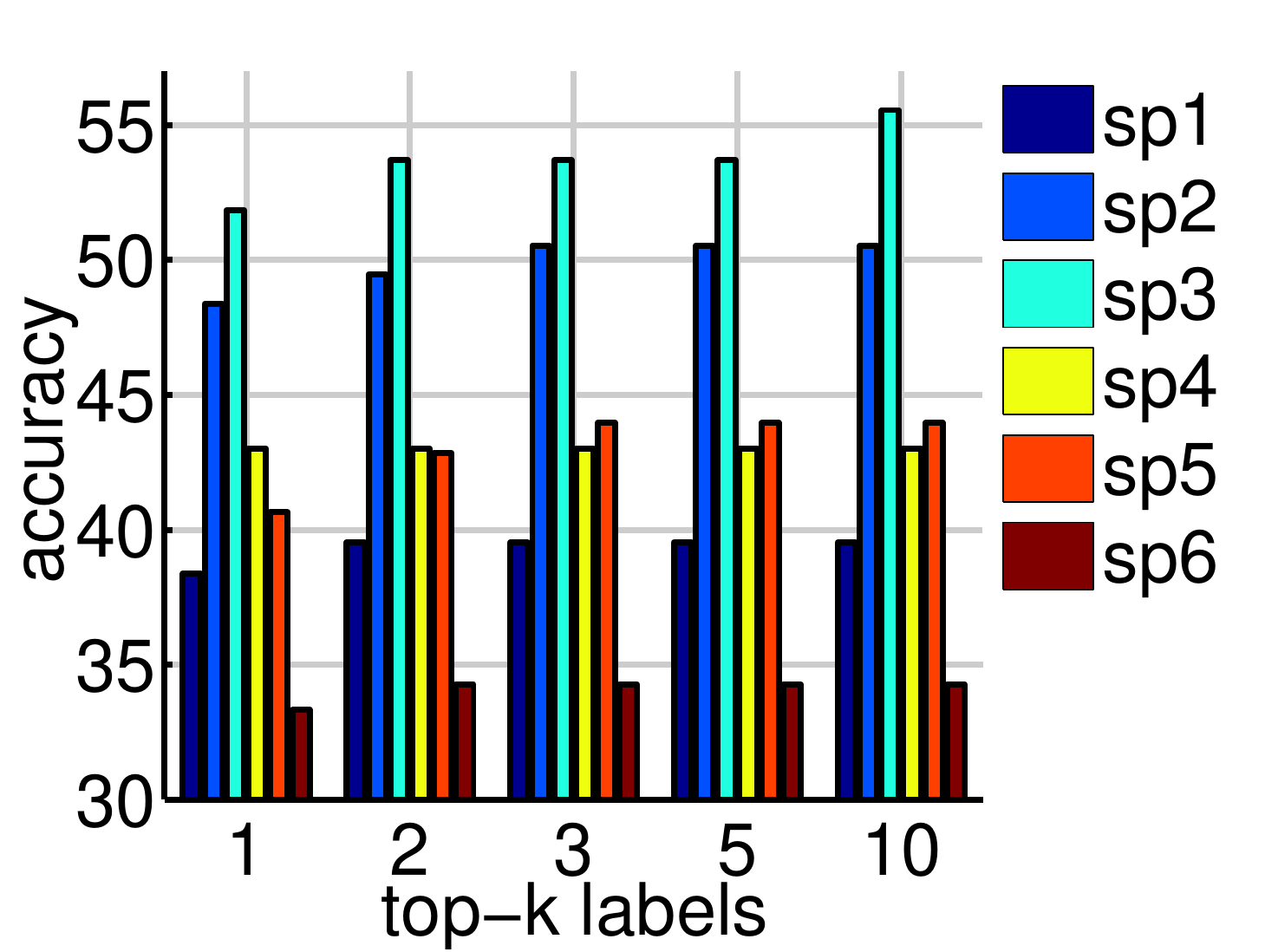}
\caption{\label{fig:rec_paint1}}
\end{subfigure}
\begin{subfigure}[b]{0.49\linewidth}
\centering\includegraphics[trim=0 0 0 0, clip=true, width=6.0cm]{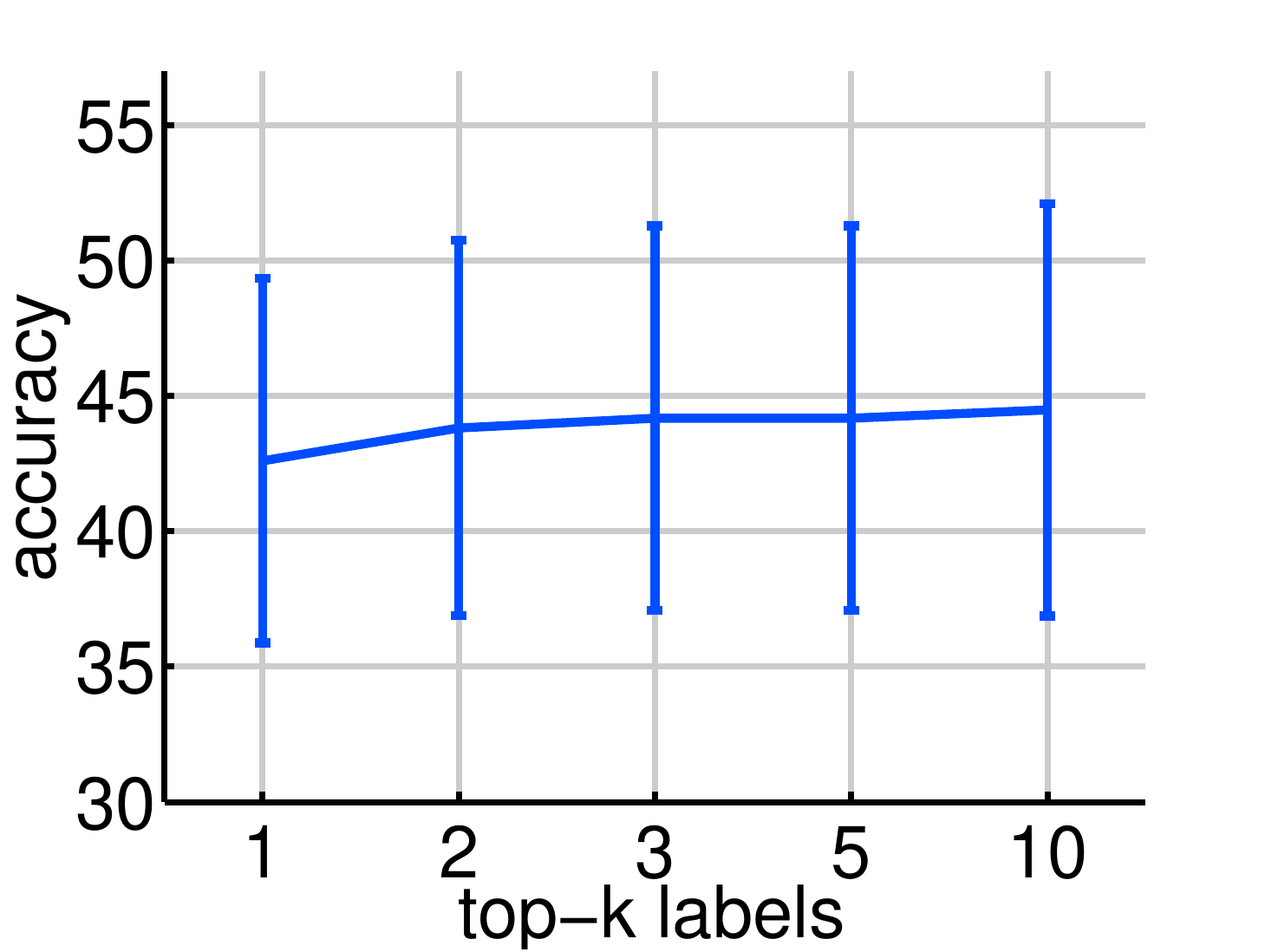}
\caption{\label{fig:rec_paint2}}
\end{subfigure}
%
\caption{The Paintings dataset. Figure \ref{fig:rec_paint1} illustrates accuracy in percents (the higher the better) for each of six testing splits -- each collected by a different volunteer. Figure~\ref{fig:rec_paint2} shows the average over the six splits as well as the standard deviation.}
\label{fig:rec_paint}
\end{figure}

Figure \ref{fig:rec_paint1} illustrates performance obtained on the Paintings dataset for each of six testing splits from six volunteers. We count prediction as a valid piece of identification if the predicted label is within {\em top-k} ground truth labels (k being a number along axis) assigned by us in the data annotation process. As demonstrated, most of the predictions point to the central pieces in images from wearable cameras; therefore, accuracy improves only marginally as the {\em top-k} value increases in the plot. For instance, split sp4 shows no variation w.r.t. the {\em top-k} value. However, splits sp3 and sp5 show close to $4\%$ variation. This can be explained by the fact that volunteers who collected the data for these two splits tended to stroll along the exhibition space away from paintings. Therefore, many images collected this way contained several paintings. Figure \ref{fig:rec_paint2} shows the average performance over the six splits. As demonstrated, due to differences in how volunteers explored the museum space and mounted wearable cameras on their clothing, the standard deviation between results varies by up to $\pm\!6.7\%$. The best performing split, sp3, scored $51.8\%$ accuracy while the worst performing split scored only $33.3\%$ accuracy. This highlights the difficulty in attaining equally good recognition rates for the data from every visitor. The average accuracy for top-1 labels obtained in this experiment is $42.6\%$, which means that exactly such a portion of all images from wearable cameras were recognized correctly.

\begin{figure}[t]
\centering\vspace*{-0.6cm}
\begin{subfigure}[b]{0.49\linewidth}
\centering\includegraphics[trim=0 0 0 0, clip=true, width=6.0cm]{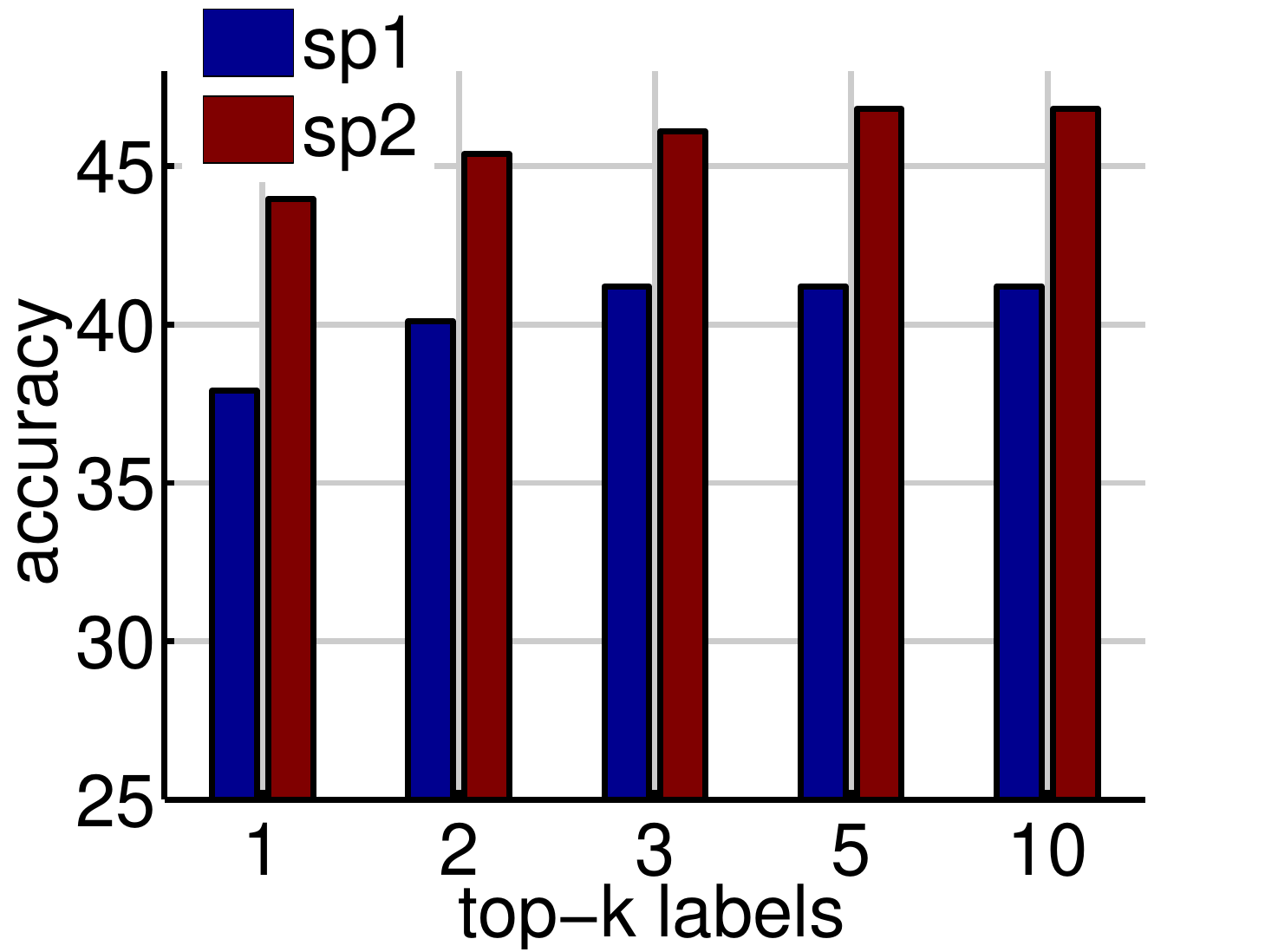}
\caption{\label{fig:rec_clock}}
\end{subfigure}
\begin{subfigure}[b]{0.49\linewidth}
\centering\includegraphics[trim=0 0 0 0, clip=true, width=6.0cm]{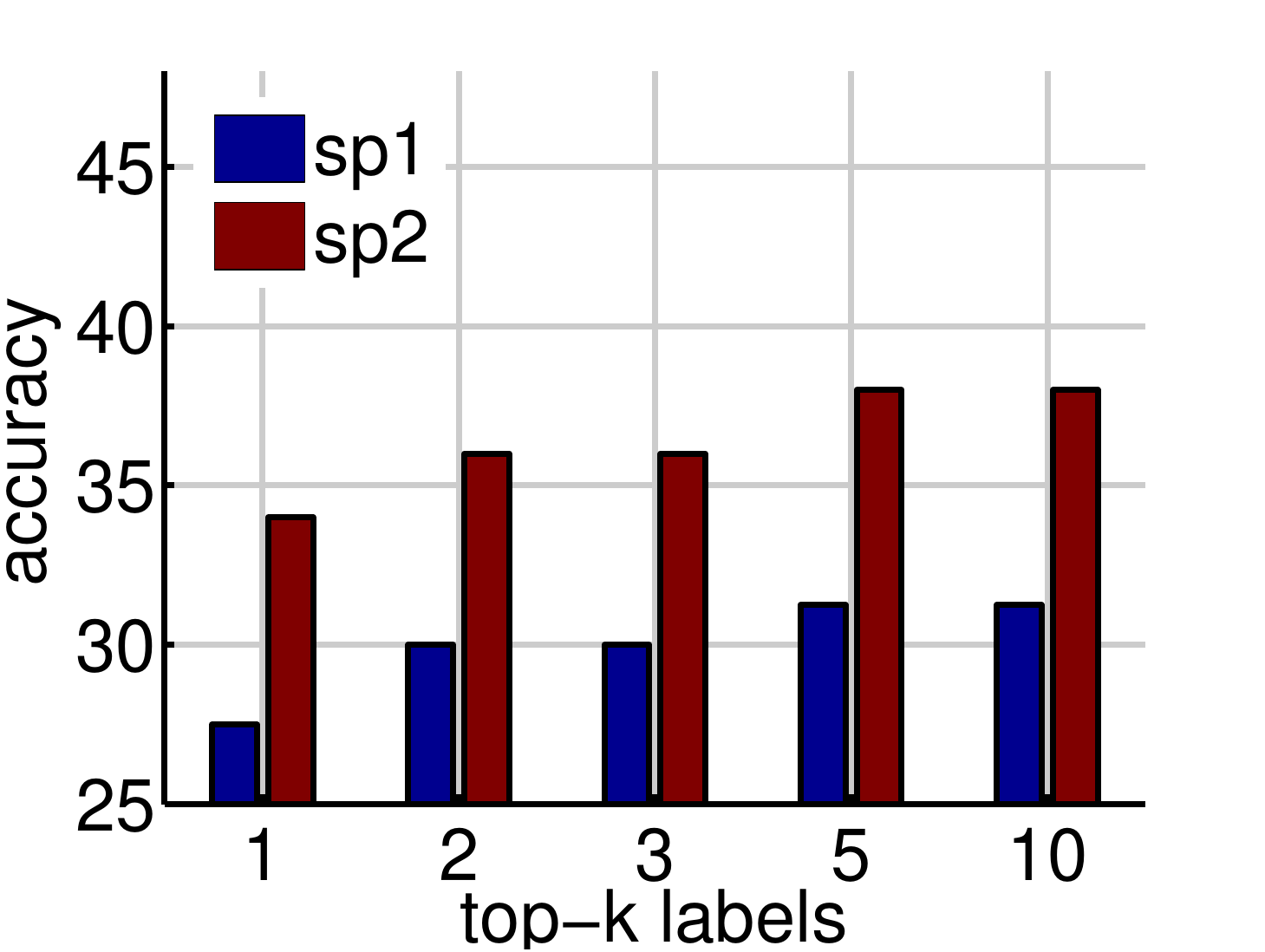}
\caption{\label{fig:rec_sculpture}}
\end{subfigure}
%
\caption{The Clocks and Sculpture datasets are evaluated in Figures \ref{fig:rec_clock} and \ref{fig:rec_sculpture}, respectively. The mean accuracy in percents is indicated by the bar plots.}\vspace{-0.3cm}
\label{fig:rec_sculpture_paint}
\end{figure}

Figure \ref{fig:rec_clock} shows performance on the Clock dataset for both testing splits. As demonstrated, recognition rates differ by $6.1\%$ between these two testing sets. We suspect this highlights a big difference in how the two volunteers explored this museum space. Another explanation is that recognition is affected by the way visitors mounted wearable cameras. However, we also note that additional ground truth labels (when multiple clocks were visible in an image) turned out to be not needed as the accuracy for larger {\em top-k} values (e.g. top-2, …,top-10) increases by up to $3.3\%$. We suspect that because clocks were located behind protective glass, visitors approached each artwork and explored it up-close. Therefore, wearable cameras were able to obtain clear, well centered pictures of most of the timepieces. The average accuracy for top-1 labels obtained in this experiment is $40.9\%$, which is slightly below the average accuracy of the Paintings dataset. We note that this dataset constitutes a contrast with the Paintings dataset. We expected that recognition of non-planar artworks behind the protective glass in a darker and more crowded environment would be a harder task; however, the need to approach these pieces helped the cameras capture their clear close-up pictures.

Figure \ref{fig:rec_sculpture} shows the performance on the Sculptures dataset for both testing splits. Firstly, we note that in some cases the difference in accuracy for top-1 vs. top-10 measure differs by up to $4\%$. We expect this is due to other sculptures present in the background. The CNN network was very likely unable to distinguish between the central object and other surrounding items. Moreover, we also expect some noise in our ground truth annotations, as sometimes it was not clear which object in an image was the central object approached by the volunteer. Lastly, we note that the average accuracy for top-1 labels obtained in this experiment is only 30.7\% which is a drop of over 10\% compared to results on the Paintings and Clocks datasets. This highlights a challenge of identifying non-planar artworks in cluttered exhibition spaces.

Because we are interested in identifying artworks that the volunteers interacted with, for each dataset, we asked one of the volunteers to approach all artworks in a given museum space. For paintings, clocks and sculptures, we were able to recognize 36, 54, and 15 distinct paintings, clocks and sculptures out of 79, 113, and 44 distinct art pieces in each exhibition space. This means that the fine-tuned CNN was able to recognize 45.6\%, 47.8\% and 34.1\% of all distinct artworks.

\section{Conclusions}

This work addresses the challenging problem of artwork identification in museum spaces. We have shown that, with state-of-the-art computer vision CNN algorithms, we are able to reliably identify up to half of the artworks that audiences interact with in various museum spaces. We found that our discussion of the challenges posed by the various types of exhibition spaces (and specific artworks) to the capturing and recognition process are indeed reflected by the quantitative results we obtained. It appears that for now, identification of paintings is perhaps the simplest task due, to their planarity. However, non-planar items such as clocks and sculptures pose a somewhat bigger challenge. Above all, this pilot study reveals that the off-the-shelf fine-tuning so popular in computer vision is perhaps still insufficient, and requires a more customized recognition algorithm. Suitable modifications may include a variation of CNN \citep{ckn} and so-called bag-of-words or domain adaptation approaches \citep{me_tensor,me_ATN,sparse_tensor_cvpr,  me_domain,me_segdes}. With just below half of the artworks identified correctly, it may be sufficient to combine the artwork identification module with a recommendation system, though the need for further improvement is clear. In the future, we plan to extend the current dataset to contain pictures from more kinds of exhibition spaces, as well as investigate new classification algorithms.


\end{document}